\pdfoutput=1

\documentclass[11pt]{article}

\usepackage[]{acl}

\usepackage{times}
\usepackage{latexsym}

\usepackage[T1]{fontenc}

\usepackage[utf8]{inputenc}

\usepackage{microtype}

\usepackage{bold-extra}   

\usepackage{microtype}    
\usepackage{graphicx}
\usepackage{multirow}
\usepackage{amsfonts}     
\usepackage{amssymb}
\usepackage{amsmath}
\usepackage{bm}           

\usepackage{caption}
\usepackage{subcaption}

\usepackage[english]{babel}
\usepackage[utf8]{inputenc}
\newcommand{\LN}{\linebreak\noindent}    
\newcommand{\MC}[3]{\multicolumn{#1}{#2}{#3}}
\DeclareMathOperator*{\argmax}{arg\,max}

\title{Enhancing Cognitive Models of Emotions with Representation Learning}

\author{Yuting Guo \\
  Computer Science \\
  Emory University \\
  Atlanta GA 30322, USA \\
  \texttt{yuting.guo@emory.edu} \\\And
  Jinho D. Choi \\
  Computer Science \\
  Emory University \\
  Atlanta GA 30322, USA \\
  \texttt{jinho.choi@emory.edu} \\}

\begin{document}
\maketitle
\begin{abstract}
We present a novel deep learning-based framework to generate embedding representations of fine-grained emotions that can be used to computationally describe psychological models of emotions.
Our framework integrates a contextualized embedding encoder with a multi-head probing model that enables to interpret dynamically learned representations optimized for an emotion classification task.
Our model is evaluated on the Empathetic Dialogue dataset and 
shows the state-of-the-art result for classifying 32 emotions.
Our layer analysis can derive an emotion graph to depict hierarchical relations among the emotions.
Our emotion representations can be used to generate an emotion wheel directly comparable to the one from Plutchik's\LN model, and also augment the values of missing emotions in the PAD emotional state model.
\end{abstract}


\section{Introduction}
\label{sec:introduction}

Emotion classification has been extensively studied\LN by many disciplines for decades \cite{spencer:1895, lazarus:1994, ekman:1999}.
Two main streams have been developed for this research: one is the discrete theory that tries to explain emotions with basic and complex categories \cite{plutchik:1980,ekman:1992,colombetti:2009}, and the other is the dimensional theory that aims to conceptualize emotions into a continuous vector space \cite{russell:1977,watson:1985,bradley:1992}.
Illustration of human emotion however is often subjective and obscure in nature, leading to a long debate among researchers about the ``correct'' way of representing emotions \cite{gendron:2009}.


Representation learning has made remarkable progress recently by building neural language models on large corpora, which have substantially improved the performance on many downstream tasks \cite{peters:2018,devlin:2019,yang:2019,liu:2019,joshi:2019}.
Encouraged by this rapid progress along with an increasing interest of interpretability in deep learning models, several studies have attempted to capture various knowledge encoded in language \cite{adi:2016,peters:2018,hewitt:2019},\LN and shown that it is possible to learn computational representations through distributional semantics for abstract concepts.
Inspired by these prior studies, we build a deep learning-based framework to generate emotion embeddings from text and assess its ability of enhancing cognitive models of emotions.
Our contributions are summarized 
as follows:\footnote{All our resources including source codes and models are available at \url{https://github.com/emorynlp/CMCL-2021}.}

\begin{itemize}
\setlength\itemsep{0em}
\item To develop a deep probing model that allows us to interpret the process of representation learning on emotion classification (Section~\ref{sec:multi-head-probing}).

\item To achieve the state-of-the-art result on the Empathetic Dialogue dataset for the classification of 32 emotions (Section~\ref{sec:experiments}).

\item To generate emotion representations that can derive an emotion graph, an emotion wheel, as well as fill the gap for unexplored emotions from existing emotion theories (Section~\ref{sec:Analysis}).
\end{itemize}

\begin{figure*}[htbp!]
\centering
\includegraphics[scale=0.4]{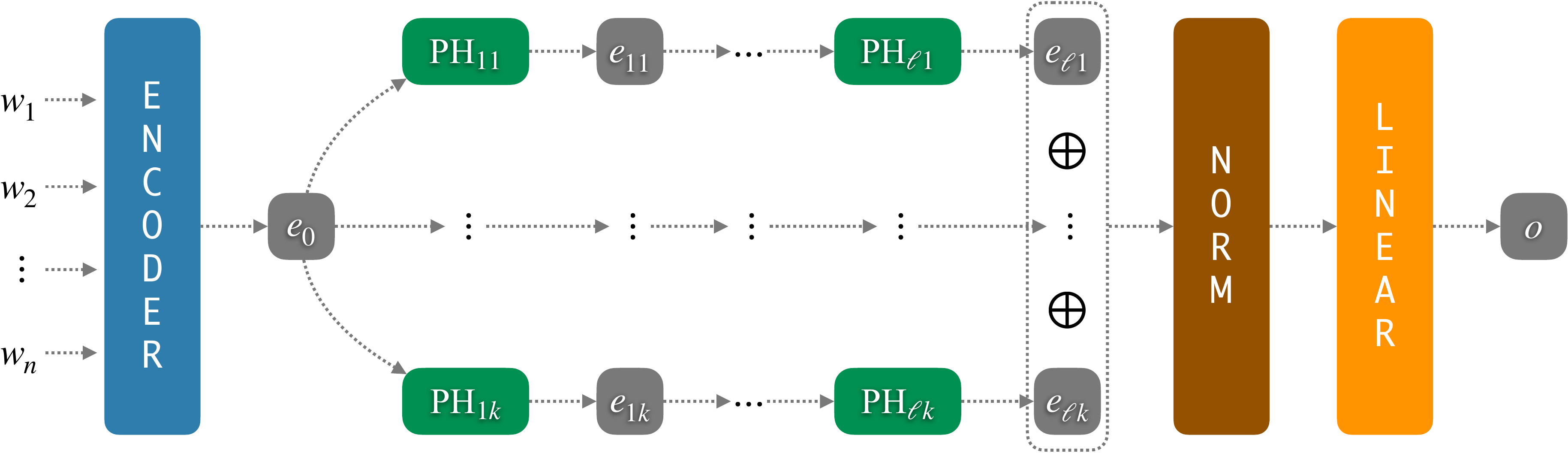}
\caption{The overview of our deep learning-based multi-head probing model.}
\label{fig:probing-model}
\vspace{-1ex}
\end{figure*}

\vspace{-2ex}
\section{Related Work}
\label{sec:related_work}


Probing models are designed to construct a probe to detect knowledge in embedding representations.
\newcite{peters:2018} used linear probes to examine phrasal information in representations learned by deep neural models on multiple NLP tasks. 
\newcite{tenney:2019} proposed an edge probing model using a span pooling to analyze syntactic and semantic relations among words through word embeddings.
\newcite{hewitt:2019} constructed a structural probe to detect the correlations among word pairs to predict their latent distances in dependency trees. 
As far as we can tell, our work is the first to generate embeddings of fine-grained emotions from text and apply them to well-established emotion theories.


\noindent NLP researchers have produced several corpora for emotion detection including FriendsED \cite{zahiri:18a}, EmoInt \cite{mohammad:2017}, EmoBank \cite{buechel:2017}, and DailyDialogs \cite{li:2017}, all of which are based on coarse-grained emotions with at most 7 categories.\LN 
For a more comprehensive analysis, we adapt the Empathetic Dialogue dataset based on fine-grained emotions with 32 categories \cite{rashkin:2018}.


\section{Multi-head Probing Model}
\label{sec:multi-head-probing}

We present a multi-head probing model allowing us to interpret how emotion embeddings are learned in deep learning models.
Figure~\ref{fig:probing-model} shows an overview of our probing model.
Let $W = \{w_1, \ldots, w_n\}$ be an input document where $w_i$ is the $i$'th token in the document.
$W$ is first fed into a contextualized embedding encoder that generates the embedding $e_0 \in \mathbb{R}^{d_0}$ representing the entire document.
The document embedding $e_0$ is then fed into multiple probing heads, $\texttt{PH}_{11}, \ldots, \texttt{PH}_{1k}$, that generate the vectors $e_{1j} \in \mathbb{R}^{d_1}$ comprising features useful for emotion classification $(j \in [1, k])$.
The probing heads in this layer are expected to capture abstract concepts (e.g., positive/negative, intense/mild).

Each vector $e_{1j}$ is fed into a sequence of probing heads where the probing head $\texttt{PH}_{ij}$ is defined $\texttt{PH}_{ij}(e_{hj}) \rightarrow e_{ij}$ $(i \in [2, \ell], j \in [1, k], h = i-1)$.
The feature vectors $e_\ell *$ from the final probing layer are expected to learn more fine-grained concepts (e.g., ashamed/embarrassed, hopeful/anticipating).
$e_\ell *$ are concatenated and normalized to $g_\ell \in \mathbb{R}^{d_\ell \cdot k}$ and fed into a linear layer that generates the output vector $o \in \mathbb{R}^{m}$ where $m$ is the total number of emotions in the training data.
It is worth mentioning that every probing sequence finds its own feature combinations.
Thus, each of $e_\ell *$ potentially represents different concepts in emotions, which allow\LN us to analyze concept compositions of these emotions empirically derived by this model.

\begin{figure*}[htbp!]
\centering
\includegraphics[width=0.8\textwidth]{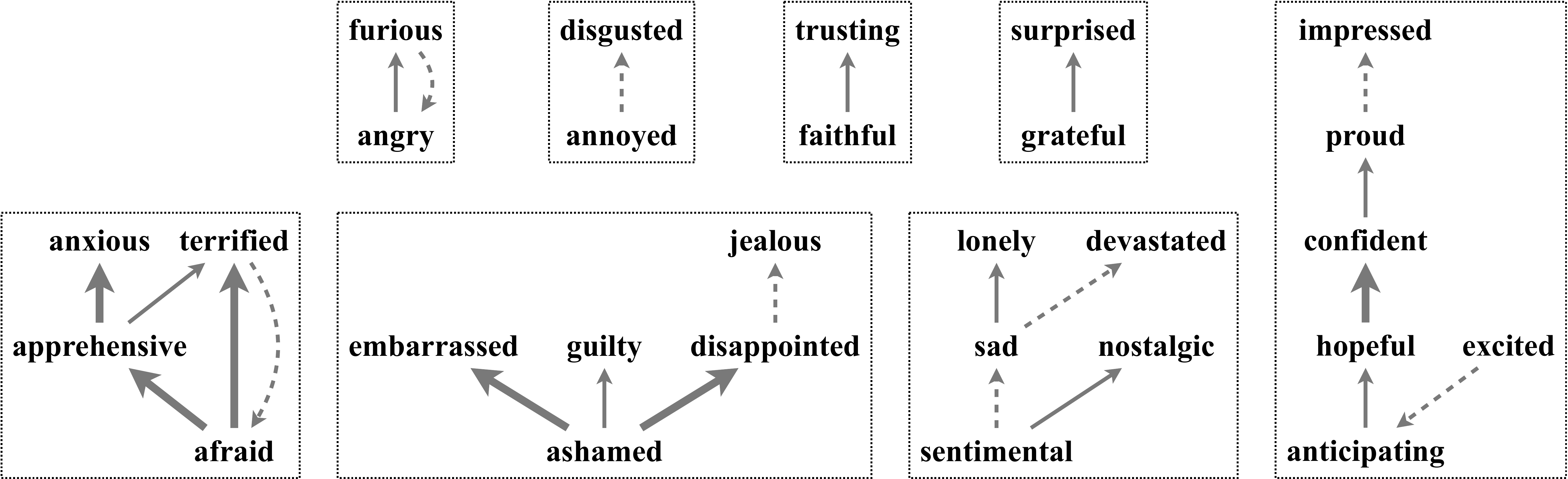}
\caption{The overview of our deep learning-based multi-head probing model.}
\label{fig:layer-analysis}
\vspace{-2ex}
\end{figure*}

\section{Experiments}
\label{sec:experiments}


\subsection{Contextualized Embedding Encoder}
\label{ssec:document-encoder}

For all experiments, BERT \cite{devlin:2019} is used as the contextualized embedding encoder for our multi-head probing model in Section~\ref{sec:multi-head-probing}. 
BERT prepends the special token \texttt{CLS} to the input document $W$ such that $W' = \{\texttt{CLS}\} \oplus W$ is fed into the \textsc{encoder} in Figure~\ref{fig:probing-model} instead, which generates the document embedding $e_0$ by applying several layers of multi-head attentions to \texttt{CLS} along with the other tokens in $W$ \cite{vaswani:2017}.\footnote{Details about the experimental settings are provided in Section~\ref{ssec:experimental-settings}.}



\subsection{Dataset}
\label{ssec:dataset}

Although several datasets are available for various types of emotion detection tasks (Section~\ref{sec:related_work}), most of them are annotated with coarse-grained labels that are not suitable to make a comprehensive analysis of emotions learned by deep learning models.

\begin{table}[htbp!]
\centering\resizebox{\columnwidth}{!}{
\begin{tabular}{c||c|c|c|c}
 & \textbf{\texttt{TRN}} & \textbf{\texttt{DEV}} & \textbf{\texttt{TST}} & \textbf{\texttt{ALL}} \\
\hline\hline
\tt C & 19,533           & 2,770            & 2,547            & 24,850 \\ 
\tt L & 18.2 ($\pm$10.4) & 19.6 ($\pm$11.4) & 23.0 ($\pm$12.5) & 18.9 ($\pm$10.8) \\
\end{tabular}}
\caption{Statistics of the Empathetic Dialogue dataset. \texttt{TRN}/\texttt{DEV}/\texttt{TST}: training/development/test set. \texttt{C}: \# of documents, \texttt{L}: average \# of tokens and its standard deviation in each document.}
\label{tbl:dataset}
\end{table}

\noindent To demonstrate the impact of our probing model, the Empathetic Dialogue dataset is selected, that is labeled with 32 emotions on $\approx$25K conversations related to daily life, each of which comes with an emotion label, a situation described in text that can reflect the emotion  (e.g., \texttt{Proud} $\rightarrow$ ``\textit{I finally got that promotion at work!''}), and a short two-party dialogue generated through MTurk that simulates a conversation about the situation \cite{rashkin:2018}.
For our experiments, only the situation parts are used as input documents. 



\subsection{Results}
\label{ssec:results}

Several multi-head probing models are developed by varying the number of probing layers and the dimension of feature vectors to find the most effective model for interpretation.
For all models, a linear layer is used for every probing head such that $\texttt{PH}_{i*}(e_{h*}) \rightarrow w\cdot e_{h*} = e_{i*}$, where $e_{h*} \in \mathbb{R}^{1 \times d_h}$, $w \in \mathbb{R}^{d_h \times d_i}$, $e_{i*} \in \mathbb{R}^{1 \times d_i}$.
The dimension of the document embedding $d_0$ is set to 768 for all models as configured by the pretrained BERT model. 

\begin{table}[htbp!]
\centering\small 
\begin{tabular}{r||c|c|c}
\bf\bm{$k$} & \bf 128:64:32 & \bf 64:32 & \bf 32 \\
\hline\hline
 2 &         56.9  ($\pm$0.4) &         57.1  ($\pm$0.5) & 56.9 ($\pm$0.5) \\
 4 &         57.5  ($\pm$0.4) &         58.1  ($\pm$0.5) & \textbf{57.8} ($\pm$0.5) \\
 8 & \textbf{57.8} ($\pm$0.8) & \textbf{58.2} ($\pm$0.5) & 57.6 ($\pm$0.1) \\
16 &         57.2  ($\pm$0.3) &         57.6  ($\pm$0.4) & 57.7 ($\pm$0.6) \\
32 &         57.2  ($\pm$0.9) &         57.3  ($\pm$0.4) & 57.5 ($\pm$0.7) \\
64 &         56.8  ($\pm$0.6) &         57.2  ($\pm$0.3) & 57.4 ($\pm$0.4) \\
\end{tabular}
\caption{Average accuracies and standard deviations on the test set. $k$: total \# of feature vectors in each layer, $i$'th \# in each column delimited by colons is the dimension of the feature vectors in the $i$'th probing layer.}
\label{tbl:results-ed}
\vspace{-2ex}
\end{table}

\noindent Table~\ref{tbl:results-ed} shows the results achieved by all models; every model is trained 3 times and the average accuracy and its standard deviation is reported.
The baseline BERT model using no probing, that is to feed $e_0$ directly into the linear layer, is also built for comparison, showing a significantly higher accuracy of 57.6\% ($\pm$0.02) than the previously reported state-of-the-art of 48\% by \newcite{rashkin:2018}.
The best result is achieved by the 2-layer probing model with 8 feature vectors, showing the accuracy of 58.2\% ($d_1 = 64, d_2 = 32, k = 8$). 

\section{Analysis}
\label{sec:Analysis}


\subsection{Layer-wise Analysis}
\label{ssec:Layer-wise Analysis}

To analyze which emotional concepts are embedded in each probing layer (Section~\ref{sec:multi-head-probing}), we train a logistic regression model on the concatenated vector of $(e_{i 1} \oplus \cdots \oplus e_{i k})$ for each layer $\ell_{i}$ with the same configuration used for the 3-layer model, 128:64:32 (Table~\ref{tbl:results-ed}), and tested on the development set.
For each pair of adjacent layers $(\ell_i, \ell_j)$ where $j = i+1$ and $1 \leq i \leq 2$, we measure the likelihood $H_{ij}(s, t)$ of those layers classifying each emotion $s$ as every other emotion $t$ as follows:
\begin{align*}
H_{ij}(s, t) & = L(s, t) - L(t, s) \\
L(e_g, e_p) &= \ell_j(e_g, e_p) - \ell_i(e_g, e_p)
\end{align*}
where $\ell_*(e_g, e_p)$ is the proportion of the documents whose gold labels are $e_g$ but predicted as $e_p$ by the model trained on the layer $\ell_*$.
If $L(s, t) > 0$, it means that the higher layer $\ell_j$ tends to predict $s$ as $t$ more than the lower layer $\ell_i$.
$L(t, s) > 0$ implies the opposite, and is used as a penalty term to get a more reliable measurement of how much the higher layer is confused $s$ for $t$ than the lower layer.

The results are illustrated in Figure~\ref{fig:layer-analysis}, where arrows pointing from one emotion $s$ to another emotion $t$ indicate $H_{ij}(s, j) \geq 2$.
The dashed arrows and thin solid arrows correspond to the confusion likelihoods of $H_{12}(s, j)$ and $H_{23}(s, j)$ respectively, and the thick solid arrows reflect the likelihoods in those two metrics.
Most emotion pairs point from coarse-grained emotions to fine-grained emotions (e.g., \textit{angry} $\rightarrow$ \textit{furious}, \textit{sentimental} $\rightarrow$ \textit{nostalgic}) except for a few pairs (\textit{excited} $\rightarrow$ \textit{anticipating}), implying that higher probing layers tend to learn more finer-grained emotions that lower layers.

\subsection{Generation of Emotion Wheel}
\label{ssec:emotion-wheel-generation}

\begin{figure}[htbp!]
    \centering
    \includegraphics[width=\columnwidth]{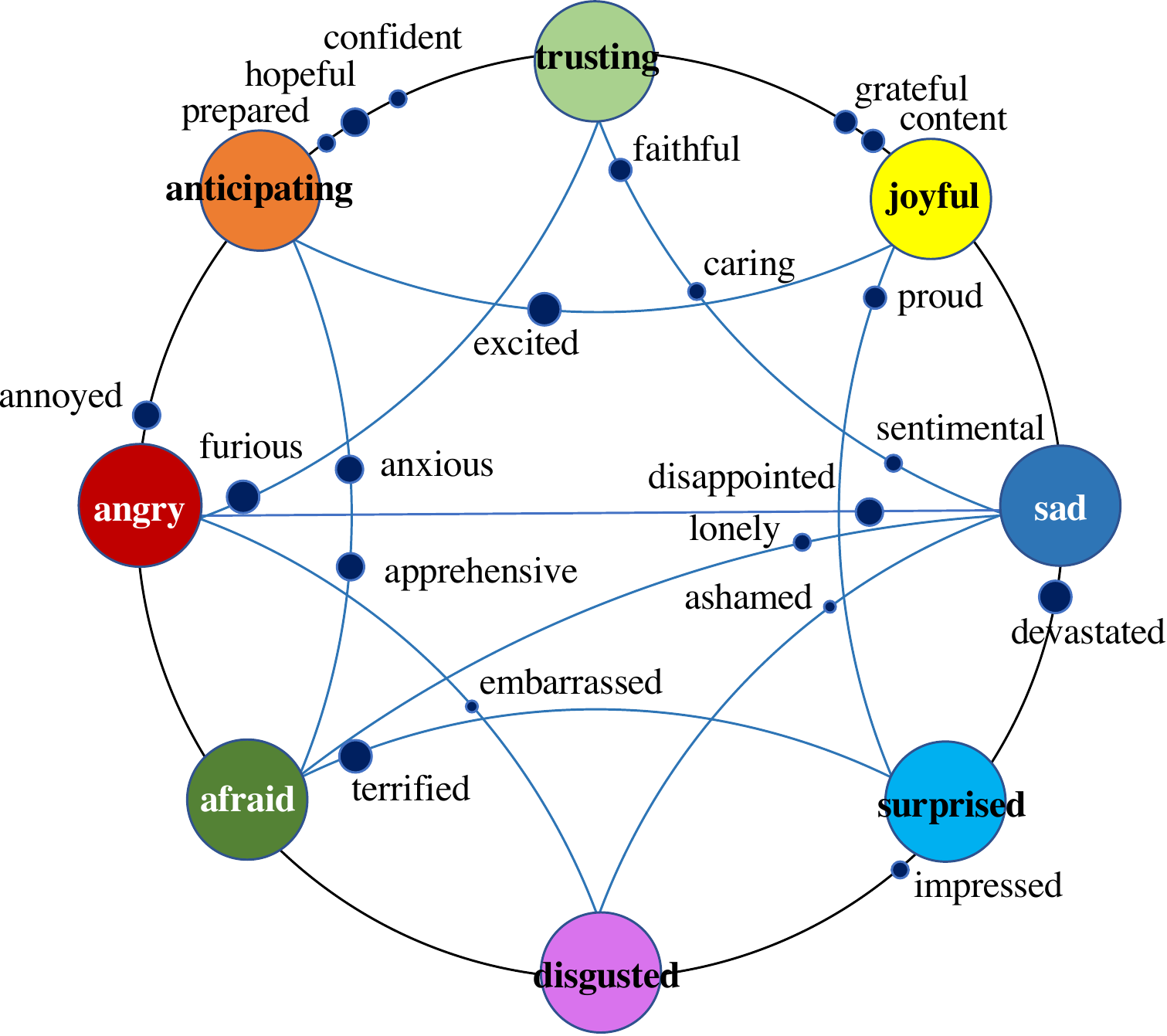}
    \caption{Emotion wheel auto-derived by our approach.}
    \label{fig:generated_emotion_wheel}
\end{figure}

\newcite{plutchik:1980} introduced the emotion wheel by selecting a reference emotion and arranging others on a circle where the angles are determined by manually assessed similarities between emotion pairs.
Inspired by this work, we derive an emotion wheel by creating emotion embeddings and representing each complex emotion as a weighted sum of two basic emotions.
Given an emotion $e$ and a set of documents $\mathcal{D}_e$ whose gold labels are $e$ in the \texttt{DEV} set, the embedding of $e$ can be derived as follows, where $g_\ell^d$ is the normalized vector in Section~\ref{sec:multi-head-probing} for $d$.\LN
\begin{equation}
r_e = \frac{1}{|\mathcal{D}_e|}\sum_{\forall d \in \mathcal{D}_e} g_\ell^d
\end{equation}
For each complex emotion $c$, its combinatory basic emotion pair $(b_i, b_j)$ and the weight $w \in [0.1, 0.9]$ are founded as follows ($r_*$ is the embedding of $b_*$):
\begin{align}
r_{i,j,w} &= w \cdot r_i + (1-w) \cdot r_j\nonumber \\
(b_i, b_j, w) &= \argmax_{\forall i, \forall j, \forall w} [\text{cosine\_sim}(r_{i,j,w}, c)]
\end{align}
Figure~\ref{fig:generated_emotion_wheel} depicts the emotion wheel auto-generated by our framework; 8 basic emotions are displayed on the outer circle and complex emotions are displayed on the edges between those basic emotions where the dot scales are proportional to the cosine\_ sims in Eq (2).\footnote{3 complex emotions whose cosine similarity scores are less than 0.1 are omitted in Figure~\ref{fig:generated_emotion_wheel}: \textit{guilty}, \textit{jealous}, \textit{nostalgic}.}
Although the only manual part in this wheel is the selection of those basic emotions from \newcite{plutchik:1980}, it is compatible to the original emotion wheel in Section~\ref{ssec:plutchk-emotion-wheel} and finds even more relations such as {\small \textit{Excited} = \textit{Anticipating} + \textit{Joyful}}, {\small \textit{Lonely} = \textit{Sad} + \textit{Afraid}}, and {\small \textit{Grateful} = \textit{Trusting} + Joyful}.

\vspace{-2.5ex}
\subsection{Augmentation of PAD Model}
\label{ssec:PAD Model Comparison}

\newcite{russell:1977} presented the PAD model suggesting that emotions can be denoted by 3 dimensions of pleasure, arousal, and dominance.
To verify whether our representations can capture emotional concepts similar to the PAD model, we train a regression model per dimension that takes the emotion embeddings from Eq (1) and learns the corresponding PAD values in Section~\ref{ssec:augment-pad} manually assessed by \newcite{russell:1977}.

\begin{figure}[htbp!]
    \centering
    \includegraphics[width=\columnwidth]{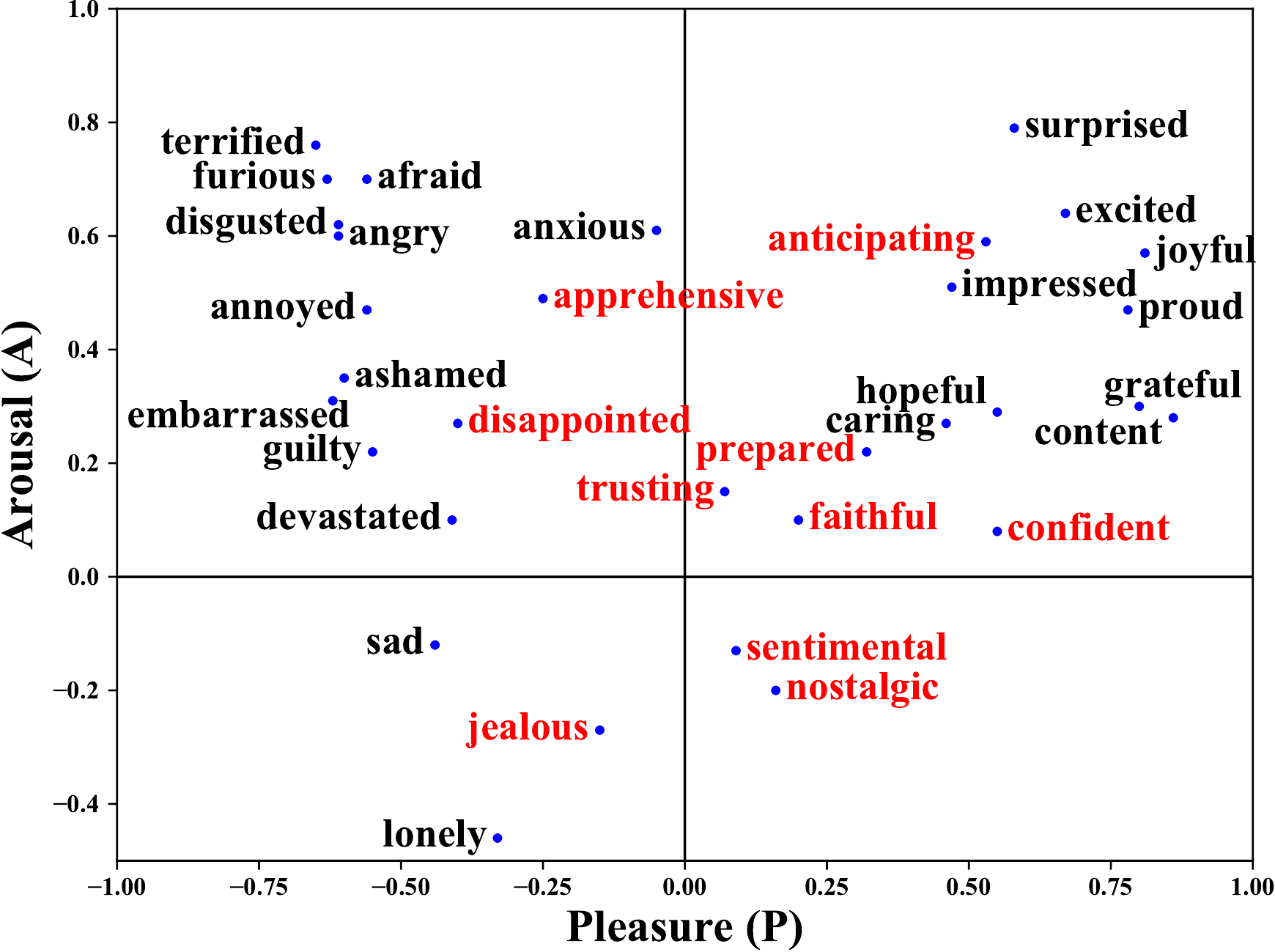}
    \caption{The 2D plot from the PAD values of 32 emotions predicted by our regression models.}
    \label{fig:pad_2d}
\end{figure}

\noindent Note that the original PAD model provides the PAD values for only 22 emotions.
Given the 3 regression models trained on those 22 emotions, we are able to predict the PAD values for the other 10 emotions missing from the original model.\footnote{Section~\ref{ssec:augment-pad} provides configurations for all three models.}
Figure~\ref{fig:pad_2d} shows the 2D plot of the PA values predicted by our regression models for Pleasure and Arousal, where the 10 emotions, whose PAD values are newly discovered by our models, are indicated with the red labels.\footnote{The 3D plot including the dominance values is in Section~\ref{ssec:augment-pad}.}
It is exciting to see that the newly discovered emotions blend well in this plot (e.g., \textit{anticipating} in between \textit{anxious} and \textit{excited}).
Similar emotions are closer in this space (e.g., \textit{sentimental} / \textit{nostalgic}, \textit{trusting} / \textit{faithful} / \textit{confident}), implying the robustness of the predicted values.
Notice that the P value\LN of \textit{nostalgic} is predicted as positive, which is understandable because \textit{nostalgic} is related to a memory with happy personal associations; thus, it is found to be positive by distributional semantics.

\vspace{-2ex}
\section{Conclusion}
\label{sec:conclusion}

This paper presents a multi-head probing model to derive emotion embeddings from neural model interpretation.
Our model is applied to an emotion detection task and shows a state-of-the-art result.\LN
These emotion embeddings can derive an emotion graph, depicting how abstract concepts are learned in neural models, and an emotion wheel and PAD values, verifying their potential of augmenting cognitive models for more diverse groups of emotions that have not been explored by cognitive theories.


\bibliography{anthology,custom}
\bibliographystyle{acl_natbib}

\clearpage\appendix
\section{Appendix}
\label{sec:appendix}


\subsection{Experimental Settings}
\label{ssec:experimental-settings}

The BERT model used in our experiment is \textit{BERT-base}, and Table~\ref{tbl:hyperparameters} shows the hyperparameters used to develop the models in Table~\ref{tbl:results-ed}.

\begin{table}[htbp!]
\centering

\begin{subtable}{\columnwidth}
\centering\small
\begin{tabular}{l|r}
    \multicolumn{1}{c|}{\bf Hyperparameter} & \multicolumn{1}{c}{\bf Value} \\
    \hline\hline
    $n$: max document length & 128 \\
    $m$: number of classes   &  32 \\
    $k$: number of feature vectors in each layer & 8 \\
    $d_{0}$: dimension of the feature vector $e_{0}$ & 768 \\
    batch size & 32 \\
    learning rate & 5e-5 \\
\end{tabular}
\caption{Shared hyperparameters.}
\label{tab:name-1}
\end{subtable}
\vspace{0.5em}

\begin{subtable}{\columnwidth}
\centering\small
\begin{tabular}{l|c|c|c}
    & \multicolumn{1}{c|}{\bf 128:64:32} & \multicolumn{1}{c|}{\bf 64:32} & \multicolumn{1}{c}{\bf 32} \\
    \hline\hline
    $l$: \# of probing layers      & 3 & 2 & 1 \\
    \hline
    $d_{1}$: dimension of $e_{1}$  & 128 & 64 & 32 \\
    $d_{2}$: dimension of $e_{2}$  &  64 & 32 &  - \\
    $d_{3}$: dimension of $e_{3}$  &  32 &  - &  - \\
\end{tabular}
\caption{Model-specific hyperparameters.}
\label{tbl:param_values}
\end{subtable}

\caption{Hyperparameter configurations for all models.}
\label{tbl:hyperparameters}
\end{table}

    
 


\subsection{Plutchik's Emotion Wheel}
\label{ssec:plutchk-emotion-wheel}

\begin{figure}[htbp!]
    \centering
    \includegraphics[width=\columnwidth]{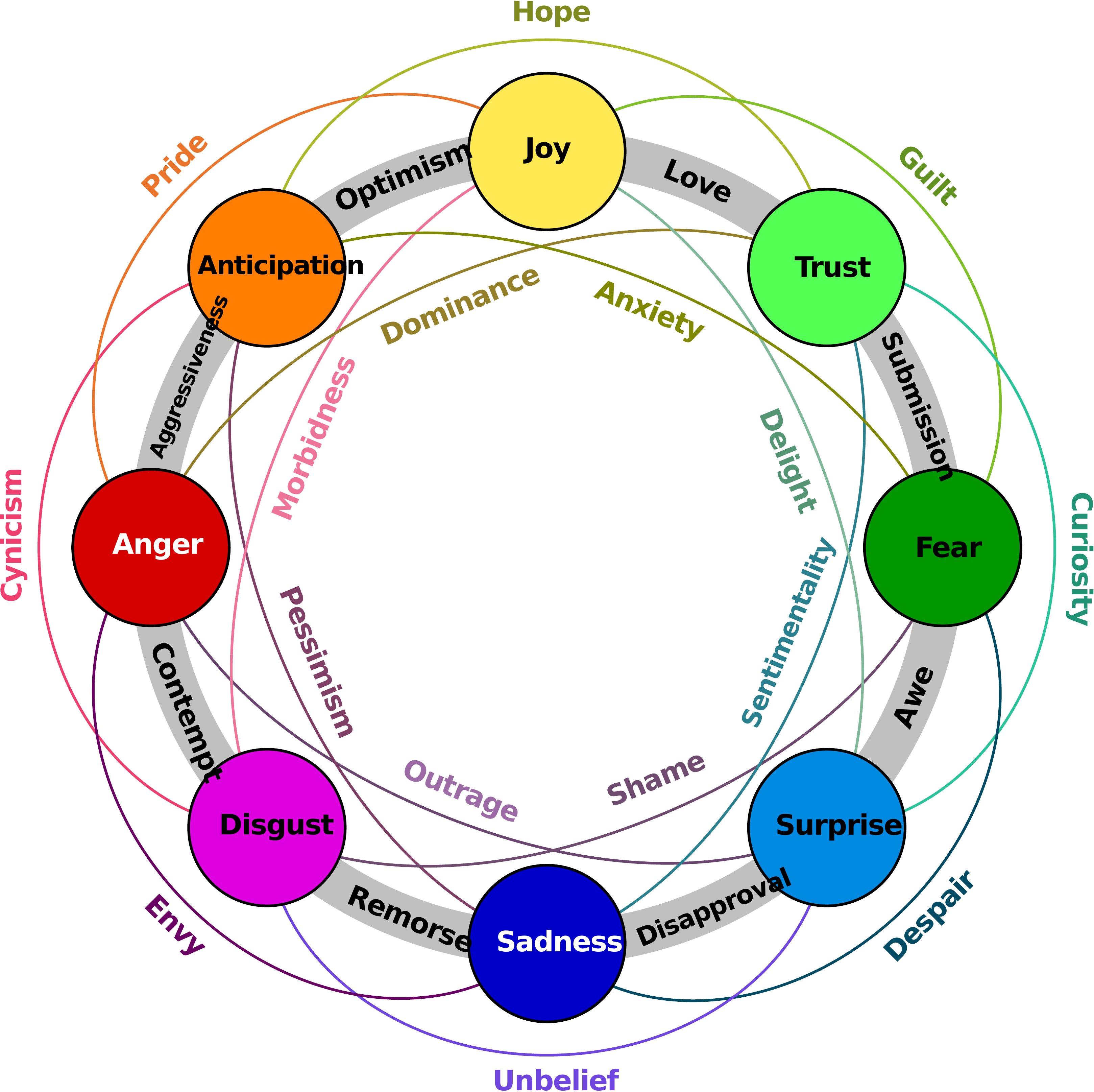}
    \caption{Emotion wheel proposed by \citet{plutchik:1980}.}
    \label{fig:plutchik-emotion-wheel}
\end{figure}

The emotion wheel described in Section \ref{ssec:emotion-wheel-generation} is inspired by \citet{plutchik:1980} which proposed the eight basic emotions that can constitute other complex emotions through various combinations shown by the emotion wheel in Figure~\ref{fig:plutchik-emotion-wheel}, where emotions displayed on the edges are the compositions of those two basic emotions.
As can be seen, our derived emotion wheel has some identical emotion relations as the Plutchik's emotion wheel such as {\small \textit{Hope} = \textit{Anticipation} + \textit{Trust}}, {\small \textit{Anxiety} = \textit{Anticipation} + \textit{Fear}}, and {\small \textit{Sentimentality} = \textit{Trust} + \textit{Sadness}}.  
It suggests the robustness of the emotion wheel derived by the proposed method in Section \ref{ssec:emotion-wheel-generation}.

\subsection{Russell and Mehrabian's PAD Model}
\label{ssec:augment-pad}

\begin{figure}[htbp!]
    \centering
    \includegraphics[scale=0.4]{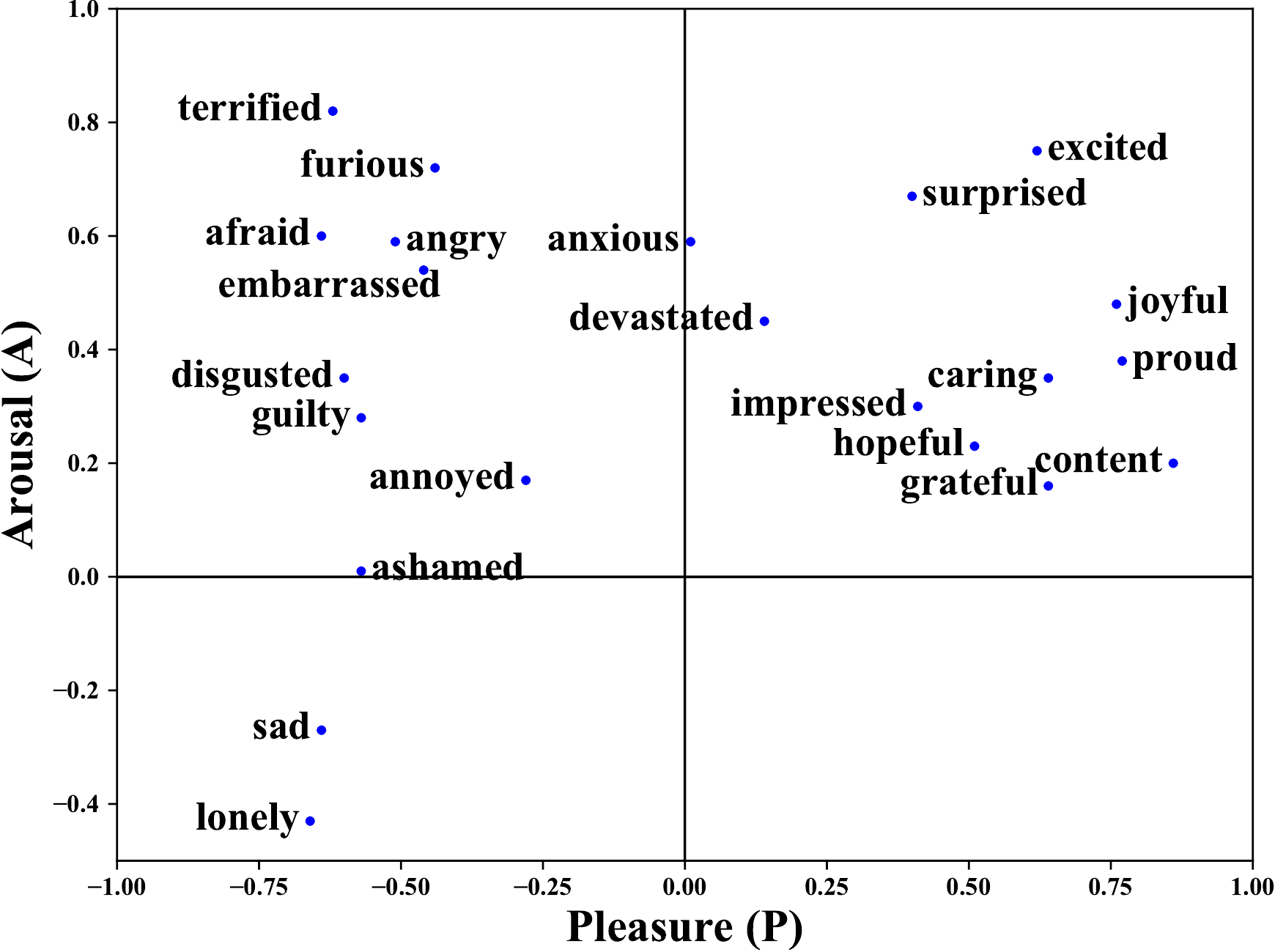}
    \caption{The 2D plot from the PAD values in Table~\ref{tbl:pad_values}.}
    \label{fig:pad_2d_original}
\end{figure}

\begin{figure}[htbp!]
    \centering
    \includegraphics[width=\columnwidth]{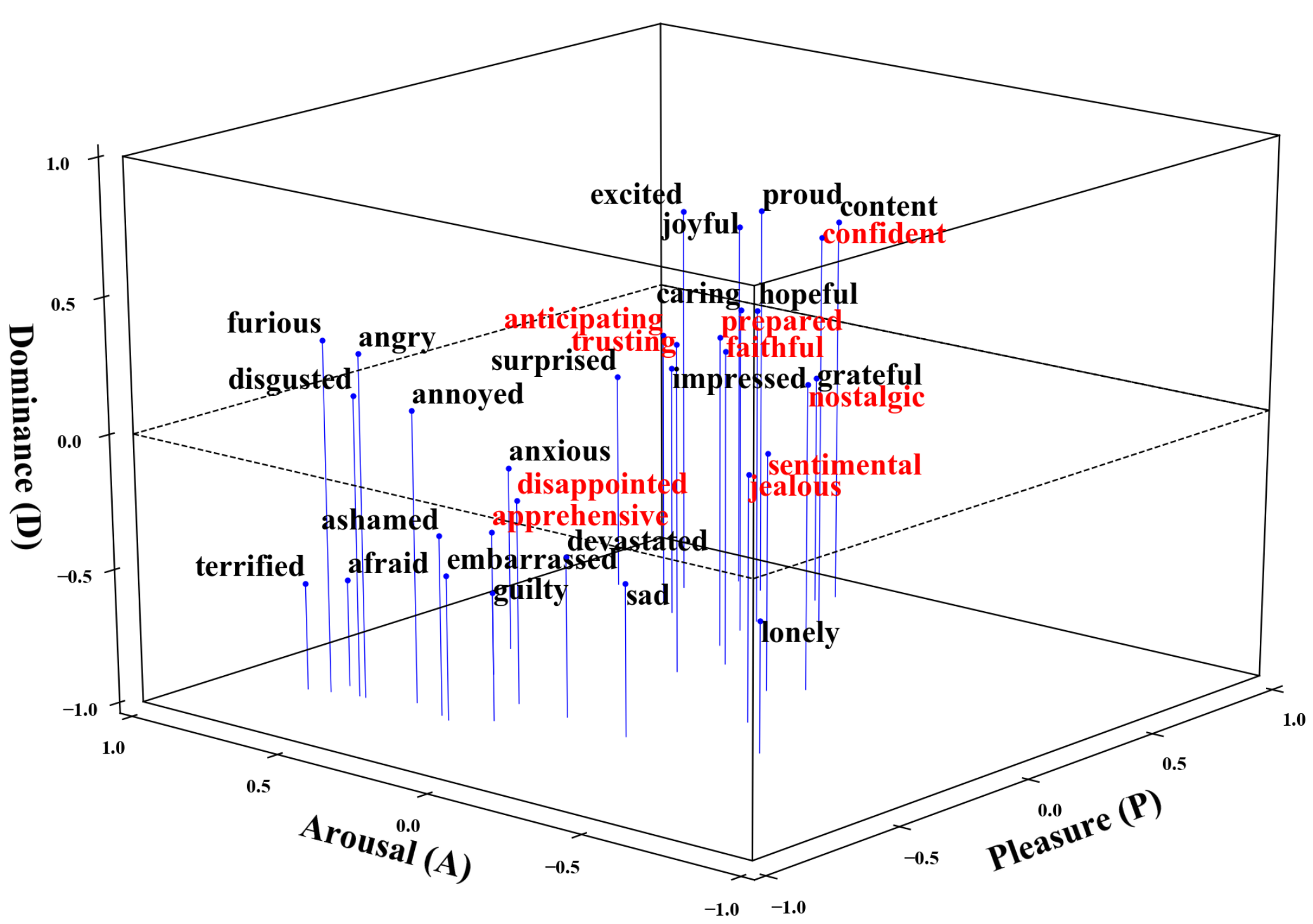}
    \caption{The 3D plot from the PAD values in Table~\ref{tbl:pad_values_predicted}.}
    \label{fig:pad_3d}
\end{figure}

All regression models in Section \ref{ssec:PAD Model Comparison} are based on 2-layer multilayer perceptron using the mean square error (MSE) loss, including a hidden layer with the \texttt{ReLU} activation and an output layer with the \texttt{Tanh}\LN activation.
The hidden layer dimension is 128, and the dropout rate is 0.3, and early stopping is applied to avoid overfitting.
The MSE losses of the three regression models to predict the Pleasure (P), Arousal (A), and Dominance (D) values are 0.028, 0.019, and 0.016, respectively.
Table \ref{tbl:pad_values} describes the original PAD values of the 22 emotions from \citet{russell:1977}, and Figure \ref{fig:pad_2d_original} shows the 2D plot from the PAD values of those 22 emotions.
Table~\ref{tbl:pad_values_predicted} describes the PAD values predicted by our regressions models, which are plotted in Figure~\ref{fig:pad_2d}.
Finally, Figure \ref{fig:pad_3d} plots those predicted PAD values in the 3D space to depict the dominance values with respect to the other two PA dimensions.
By comparing the PAD values of 22 emotions in Table \ref{tbl:pad_values} and Table \ref{tbl:pad_values_predicted}, most of the predicted values are close to their gold values.
Also, we can observe that the predicted values of some newly discovered emotions are consistent with our perception of emotions.
For example, \textit{Anticipating} is very close to \textit{Hope} in terms of pleasure but with higher intensity.
\begin{table}[htbp!]
    \centering\small
    \begin{tabular}{c||c|c|c}
    \MC{1}{c||} {\bf Emotion} & {\bf Pleasure} & {\bf Arousal} & {\bf Dominance} \\ \hline\hline
        afraid & -0.64 & 0.6 & -0.43 \\
        angry & -0.51 & 0.59 & 0.25 \\
        annoyed & -0.28 & 0.17 & 0.04 \\
        anxious & 0.01 & 0.59 & -0.15 \\
        ashamed & -0.57 & 0.01 & -0.34 \\
        caring & 0.64 & 0.35 & 0.24 \\
        content & 0.86 & 0.2 & 0.62 \\
        devastated & 0.14 & 0.45 & -0.24 \\
        disgusted & -0.6 & 0.35 & 0.11 \\
        embarrassed & -0.46 & 0.54 & -0.24 \\
        excited & 0.62 & 0.75 & 0.38 \\
        furious & -0.44 & 0.72 & 0.32 \\
        grateful & 0.64 & 0.16 & -0.21 \\
        guilty & -0.57 & 0.28 & -0.34 \\
        hopeful & 0.51 & 0.23 & 0.14 \\
        impressed & 0.41 & 0.3 & -0.32 \\
        joyful & 0.76 & 0.48 & 0.35 \\
        lonely & -0.66 & -0.43 & -0.32 \\
        proud & 0.77 & 0.38 & 0.65 \\
        sad & -0.64 & -0.27 & -0.33 \\
        surprised & 0.4 & 0.67 & -0.13 \\
        terrified & -0.62 & 0.82 & -0.43 \\
    \end{tabular}
    \caption{The original PAD values of 22 emotions provided by \citet{russell:1977}.}
    \label{tbl:pad_values}
\end{table}

\begin{table}[htbp!]
    \centering\small
    \begin{tabular}{c||c|c|c}
    \MC{1}{c||} {\bf Emotion} & {\bf Pleasure} & {\bf Arousal} & {\bf Dominance} \\ \hline\hline
        afraid & -0.56 & 0.7 & -0.6 \\
        angry & -0.61 & 0.6 & 0.28 \\
        annoyed & -0.56 & 0.47 & 0.09 \\
    \bf anticipating & \bf 0.53 & \bf 0.59 & \bf 0.03 \\
        anxious & -0.05 & 0.61 & -0.31 \\
    \bf apprehensive & \bf -0.25 & \bf 0.49 & \bf -0.46 \\
        ashamed & -0.6 & 0.35 & -0.33 \\
        caring & 0.46 & 0.27 & 0.22 \\
    \bf confident & \bf 0.55 & \bf 0.08 & \bf 0.51 \\
        content & 0.86 & 0.28 & 0.44 \\
        devastated & -0.41 & 0.1 & -0.4 \\
    \bf disappointed & \bf -0.4 & \bf 0.27 & \bf -0.24 \\
        disgusted & -0.61 & 0.62 & 0.12 \\
        embarrassed & -0.62 & 0.31 & -0.46 \\
        excited & 0.67 & 0.64 & 0.45 \\
    \bf faithful & \bf 0.2 & \bf 0.1 & \bf 0.18 \\
        furious & -0.63 & 0.7 & 0.31 \\
        grateful & 0.8 & 0.3 & -0.14 \\
        guilty & -0.55 & 0.22 & -0.52 \\
        hopeful & 0.55 & 0.29 & 0.19 \\
        impressed & 0.47 & 0.51 & -0.06 \\
    \bf jealous & \bf -0.15 & \bf -0.27 & \bf -0.08 \\
        joyful & 0.81 & 0.57 & 0.37 \\
        lonely & -0.33 & -0.46 & -0.51 \\
    \bf nostalgic & \bf 0.16 & \bf -0.2 & \bf 0.14 \\
    \bf prepared & \bf 0.32 & \bf 0.22 & \bf 0.17 \\
        proud & 0.78 & 0.47 & 0.46 \\
        sad & -0.44 & -0.12 & -0.43 \\
    \bf sentimental & \bf 0.09 & \bf -0.13 & \bf -0.11 \\
        surprised & 0.58 & 0.79 & -0.19 \\
        terrified & -0.65 & 0.76 & -0.6 \\
    \bf trusting & \bf 0.07 & \bf 0.15 & \bf 0.23 \\
    \end{tabular}
    \caption{The PAD values of 32 emotions predicted by our regression models. The 10 emotions that are missing from the original work in Table~\ref{tbl:pad_values} are indicated with bold font.}
    \label{tbl:pad_values_predicted}
\end{table}

\subsection{Combinatory Emotions Details}
\label{ssec:combinatory-emotions-details}

In Section \ref{ssec:emotion-wheel-generation}, we propose a framework to find the combinatory basic emotion pairs for each complex emotion by calculating a weighted sum vector of two basic emotion embeddings.
Table \ref{tbl:combinations} lists the basis emotion pairs, weights, and cosine similarity for 24 complex emotions derived by our framework.
The weight indicates how much each basic emotion in the pair contributes to the complex emotion and can be interpreted in a proportional manner.
For example, \textit{Annoyed} can be composed of 90\% \textit{Angry} and 10\% \textit{Anticipating}.

\begin{table}[htbp!]
    \centering\small
    \begin{tabular}{c||c|c|c|c}
    \MC{1}{c||} {$\bm{c}$} & {$\bm{b_{i}}$} & {$\bm{b_{j}}$} & {$\bm{w}$} & {\bf cos} \\ \hline\hline
        annoyed & angry & anticipating & 0.9 & 0.80 \\ \hline
        anxious & anticipating & afraid & 0.5 & 0.79 \\ \hline
        apprehensive & anticipating & afraid & 0.3 & 0.76 \\ \hline
        ashamed & sad & disgusted & 0.6 & 0.17 \\ \hline
        caring & trusting & sad & 0.5 & 0.28 \\ \hline
        confident & anticipating & trusting & 0.5 & 0.31 \\ \hline
        content & joyful & trusting & 0.9 & 0.63 \\ \hline
        devastated & surprised & sad & 0.1 & 0.93 \\ \hline
        disappointed & sad & angry & 0.7 & 0.64 \\ \hline
        embarrassed & disgusted & angry & 0.5 & 0.13 \\ \hline
        excited & anticipating & joyful & 0.5 & 0.95 \\ \hline
        faithful & trusting & sad & 0.9 & 0.59 \\ \hline
        furious & angry & trusting & 0.9 & 0.98 \\ \hline
        grateful & joyful & trusting & 0.8 & 0.56 \\ \hline
        guilty & trusting & sad & 0.1 & 0.07 \\ \hline
        hopeful & anticipating & trusting & 0.8 & 0.67 \\ \hline
        impressed & surprised & disgusted & 0.9 & 0.40 \\ \hline
        jealous & disgusted & angry & 0.3 & 0.02 \\ \hline
        lonely & afraid & sad & 0.2 & 0.33 \\ \hline
        nostalgic & anticipating & joyful & 0.1 & 0.04 \\ \hline
        prepared & anticipating & trusting & 0.9 & 0.31 \\ \hline
        proud & joyful & surprised & 0.9 & 0.45 \\ \hline
        sentimental & trusting & sad & 0.1 & 0.33 \\ \hline
        terrified & afraid & surprised & 0.9 & 0.98 \\ \hline
    \end{tabular}
    \caption{The combinatory basic emotion pairs for each complex emotion.}
    \label{tbl:combinations}
\end{table}

\end{document}